\documentclass{article}

\usepackage{microtype}
\usepackage{graphicx}
\usepackage{subcaption}
\usepackage{booktabs} 

\usepackage{multirow}
\usepackage{pifont}
\usepackage{tabularx}
\newcolumntype{Y}{>{\RaggedRight\arraybackslash}X}
\usepackage{ragged2e}
\usepackage{xcolor}
\usepackage{hyperref}
\definecolor{myCustomGreen}{RGB}{0, 176, 80}

\definecolor{myCustomBlue}{RGB}{0, 176, 240}

\usepackage[preprint]{icml2026}

\usepackage{amsmath}
\usepackage{amssymb}
\usepackage{mathtools}
\usepackage{amsthm}

\usepackage[capitalize,noabbrev]{cleveref}

\theoremstyle{plain}

\theoremstyle{definition}

\theoremstyle{remark}

\usepackage[textsize=tiny]{todonotes}

\icmltitlerunning{CROP: Expert-Aligned Image Cropping via Compositional Reasoning and Optimizing Preference}

\begin{document}

\twocolumn[
  \icmltitle{CROP: Expert-Aligned Image Cropping via Compositional Reasoning and Optimizing Preference}



  \icmlsetsymbol{equal}{*}

  \begin{icmlauthorlist}
    \icmlauthor{Zhitong Dong}{seu,lab}
    \icmlauthor{Chao Li}{comp}
    \icmlauthor{Jie Yu}{seu,lab}
    \icmlauthor{Hao Chen}{seu,lab}
  \end{icmlauthorlist}

  \icmlaffiliation{seu}{Southeast University}
  \icmlaffiliation{lab}{Key Laboratory of New Generation Artificial Intelligence Technology}
  \icmlaffiliation{comp}{Alibaba Group}

  \icmlcorrespondingauthor{Hao Chen}{haochen303@seu.edu.cn}

  \icmlkeywords{Aesthetic Image Cropping, Vision-Language Models, Direct Preference Optimization}

  \vskip 0.3in
]



\printAffiliationsAndNotice{}  

\begin{abstract}
Aesthetic image cropping aims to enhance the aesthetic quality of an image by improving its composition through spatial cropping. 
Previous methods often rely on saliency prediction or retrieval augmentation, ignoring the task's core requirement: a deep understanding of composition and aesthetics. Consequently, saliency-based methods struggle to make compositional trade-offs in complex scenes, while retrieval-based methods blindly refer to similar cases, lacking adaptive reasoning for unique scenes.
Both approaches fail to align their automated cropping results with those of human experts.
To address the above issues, we propose a novel paradigm that reformulates aesthetic cropping as a multimodal reasoning task, aiming to activate the VLM's analytical and comprehension capabilities in aesthetics.
We design a \textbf{C}ompositional \textbf{R}easoning and \textbf{O}ptimizing \textbf{P}reference method (CROP) that directs the VLM to think like a professional photographer. It deconstructs a complex and subjective aesthetic problem into an ``analysis-proposal-decision" process, reasoning step by step through the analysis of scene elements and compositional principles.
Meanwhile, our expert preference alignment module makes the model’s decision consistent with human expert aesthetics.
Extensive experiments across multiple datasets validate our method's superiority and component effectiveness.
\end{abstract}

\section{Introduction}
\label{intro}
\begin{figure}
	\centering
	\includegraphics[width=0.90\linewidth]{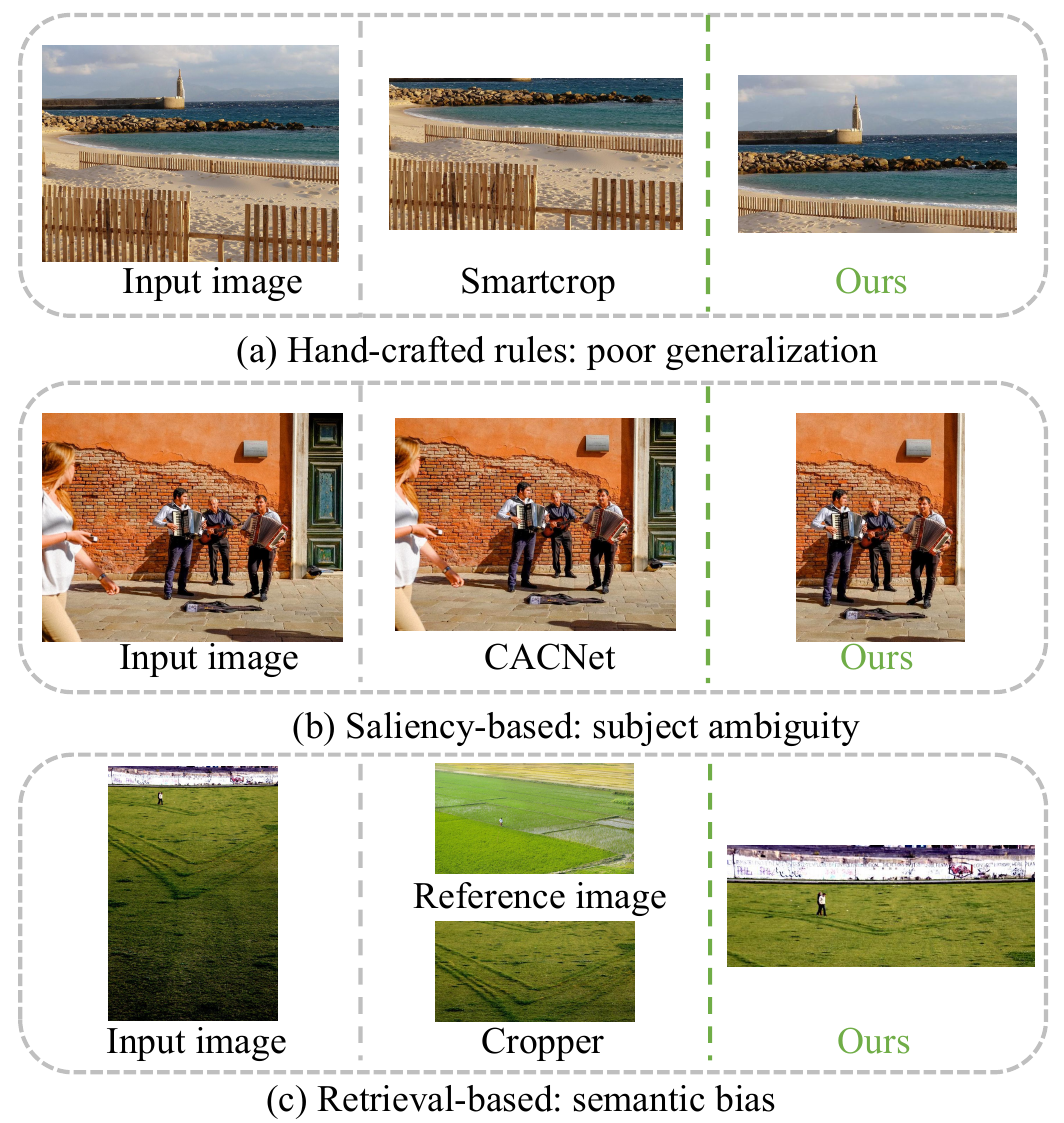}
    \vskip -0.1cm
    \caption{Smartcrop~\cite{smartcrop}, CACNet~\cite{hong2021composing}, and Cropper~\cite{lee2025cropper} represent three typical image cropping paradigms, each facing distinct limitations.}
    \label{fig:example}
\end{figure}
\begin{figure*}
	\centering
	\includegraphics[width=0.9\linewidth]{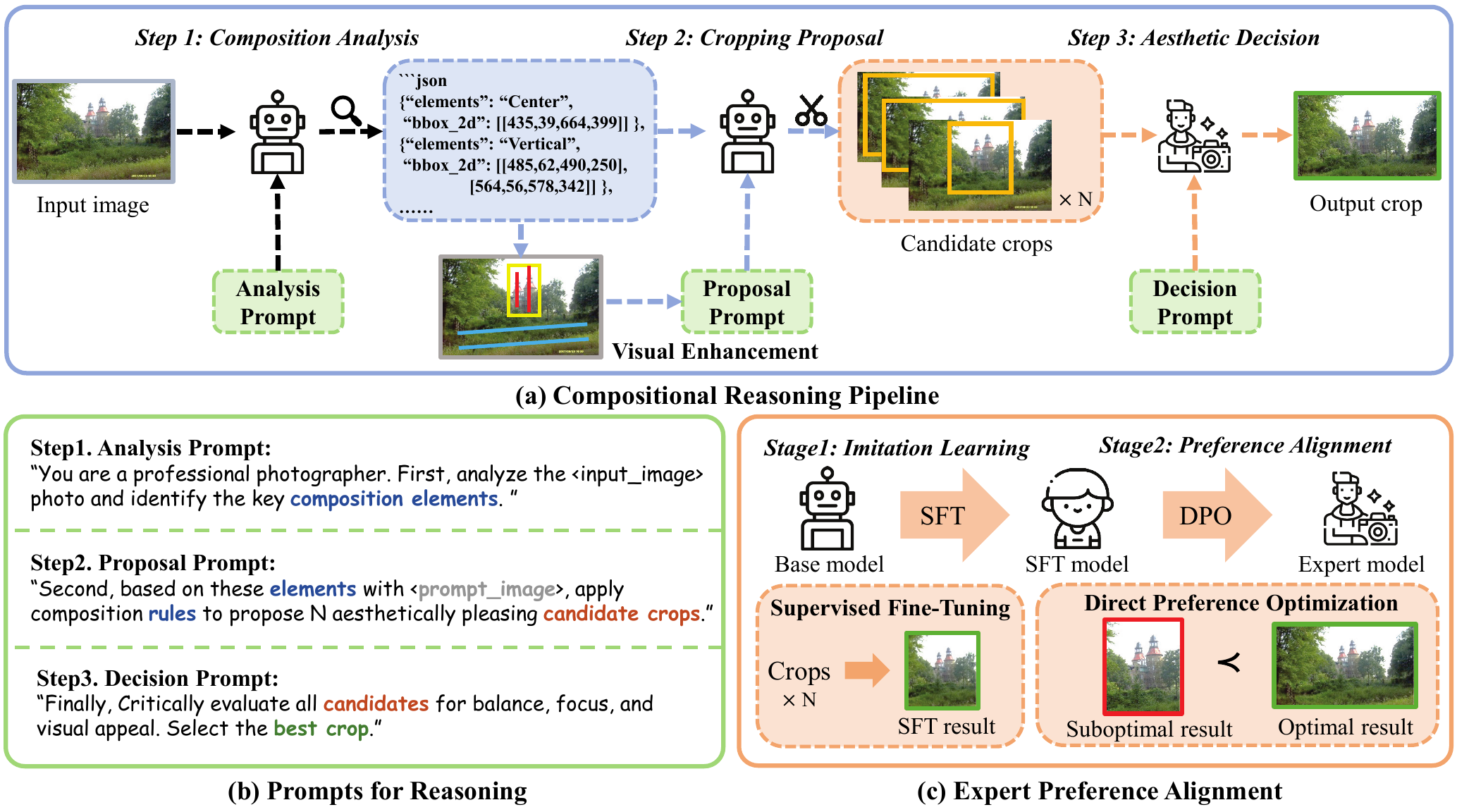}
    \caption{The overall architecture of our proposed CROP. The VLM first analyzes the key compositional elements of the input image and returns the results in JSON. Based on this analysis, CROP generates a vision-augmented prompt image and feed it back into the VLM. Leveraging both textual and visual cues, the VLM performs a second reasoning pass to propose candidate crops, which serve as inputs to the aesthetic decision module. The aesthetic decision module uses direct preference optimization to align with professional photography experts and produces the final crop. See supplementary materials for the complete prompts.}
	\label{fig:model}                                                                                             
\end{figure*}
With the rapid development of smartphone photography hardware, mobile photography has become more popular. However, good equipment alone is not enough because what matters even more is the eye behind the camera, which can be challenging for non-professionals.
Aesthetic image cropping focuses on improving the composition and visual appeal of an ordinary image through spatial cropping. Compared with other image processing tasks, this task has unique photographic and aesthetic properties.

The core challenge of aesthetic image cropping lies in how to align automated cropping results with manual works by human experts.
Traditional methods~\cite{cheng2010learning, FLMS_mm14, zhang2013weakly, nishiyama2009sensation} rely on hand-crafted features, which lack generalization and can only provide suggestions from a single perspective.
As shown in~\cref{fig:example} (a), Smartcrop~\cite{smartcrop} only focuses on the foreground, and its rigid rules lead to a lack of aesthetic appeal in the resulting composition.
Saliency-based methods~\cite{chen2017learning, li2018a2, wei2018good, zeng2020grid} typically focus on evaluating visual saliency rather than understanding scene relationships, making it difficult to achieve true compositional balance.
As shown in~\cref{fig:example} (b), when multiple competing focal points (woman and band) exist within the scene, CACNet fails to make a clear decision and instead produces a mediocre cropping result.
Recent approaches~\cite{lee2025cropper, zhang2025procrop} retrieve visually similar images from a database, which serve as learning samples to guide the cropping decision. However, the performance of such methods heavily depends on the database’s coverage and retrieval accuracy.
For example, as shown in~\cref{fig:example} (c), Cropper~\cite{lee2025cropper} performs retrieval using cosine similarity, which primarily captures semantic proximity (people walking on the grass) rather than correspondence in compositional structure. As a result, the model retrieves reference images that are semantically similar but compositionally different, thereby providing wrong guidance for cropping.

Aesthetic image cropping is inherently a multi-stage reasoning process. A photographer first surveys the scene layout, identifies key compositional elements, and then applies photographic principles to make the framing decision.
Previous methods, whether based on saliency~\cite{hong2021composing, jia2022rethinking, wang2023image, zhu2025image} or retrieval~\cite{lee2025cropper, zhang2025procrop}, rely on biased, single-step priors and fundamentally fail to emulate this step-by-step analytical process.
To address this gap, we propose a novel paradigm for image cropping that reformulates the task as a multimodal reasoning problem.
Existing VLMs struggle to effectively execute this task via instruction following.
Our core idea is to guide the VLM to learn a photographer’s thought process by first analyzing the scene’s compositional elements and then reasoning toward the optimal cropping decision.

To this end, we design a \textbf{C}ompositional \textbf{R}easoning \textbf{P}ipeline (CRP), as illustrated in~\cref{fig:model} (a). It directs the VLM to think like a professional photographer, deconstructing a complex and subjective aesthetic problem into an ``analysis-proposal-decision" process, and reasoning step by step through the analysis of scene elements and compositional principles.
Specifically, the VLM first performs (i) analysis on compositional elements, followed by (ii) proposal generation grounded in photographic rules. Since standard VLMs often overlook spatial context in favor of semantics, we reinforce visual signals through explicit annotations. Finally, the model makes an aesthetic (iii) decision on the candidates to yield the optimal crop.

Standard Supervised Fine-Tuning (SFT) alone proves inadequate to address this complex visual reasoning task. We argue that aesthetic judgment is inherently comparative—relying on ranking compositional differences—rather than a binary classification of `right' versus `wrong'. To address this, we introduce \textbf{E}xpert \textbf{P}reference \textbf{A}lignment (EPA), a comprehensive training framework that integrates SFT with Direct Preference Optimization (DPO)~\cite{rafailov2023direct}. While the SFT stage instills preliminary aesthetic capabilities, the subsequent DPO stage aligns the model's decisions with expert preferences. This framework strengthens the VLM's internal aesthetic reasoning, eliminating the need for external scorers.

After training, our method operates as a fully automated, standalone pipeline that requires no human intervention or external dependencies, ensuring efficient deployment.

Our contributions can be summarized as follows:
\begin{itemize}
\vskip -0.1cm
    \item We propose a novel paradigm for image cropping, which reformulates the task as a multimodal reasoning problem through the compositional reasoning pipeline.
    \item We introduce expert preference alignment, a two-stage training framework that aligns the VLM’s cognition with human expert aesthetics.
    \item We conduct extensive quantitative and qualitative experiments across multiple datasets to validate the superiority and generalizability of our method. Comprehensive ablation and parameter sensitivity analyses further demonstrate the effectiveness of each component.
\end{itemize}

\section{Related Work}

\subsection{Aesthetic Image Cropping}
Early approaches~\cite{cheng2010learning, FLMS_mm14, zhang2013weakly, nishiyama2009sensation, smartcrop} relying on hand-crafted features generalize poorly to complex scenes.
Recently, deep learning methods have dominated aesthetic cropping~\cite{li2018a2, wei2018good, zeng2020grid, li2020composing, jia2022rethinking, wang2023image, PanCZL0Z23, S2CNet, GenCrop, PanCWWC24}. VFN~\cite{chen2017learning} formulates the task as a viewpoint search, simulating a photographer who iteratively moves the camera.
However, this approach lacks explicit modeling of aesthetics. CACNet~\cite{hong2021composing} explicitly encodes photographic composition rules, but employing them merely to influence the cropping weights undermines interpretability.

A major challenge for image cropping is the scarcity of annotated data, as aesthetic judgments are subjective and often require photographic expertise. Datasets such as GAICD~\cite{zeng2019reliable, zeng2020grid}, FCDB~\cite{chen2017quantitative}, and FLMS~\cite{FLMS_mm14} provide numerous crop boxes and pairwise ranking annotations for crops.
To leverage these costly datasets, recent methods~\cite{lee2025cropper, zhang2025procrop} treat them as an external database and adopt retrieval-augmented learning~\cite{borgeaud2022improving, guu2020retrieval} to fetch similar exemplars during inference. However, this dependency incurs additional memory and runtime overhead, and constrains cropping diversity.
Although some methods~\cite{lee2025cropper, PhotoFramer} explore the cropping capabilities of VLMs via in-context learning or generative paradigms, leaving the native aesthetic reasoning of VLMs underexplored.
\subsection{Preference Alignment}
While pre-training and fine-tuning provide large models with broad knowledge and conversational ability, additional preference alignment is required to adapt their outputs to specific tasks or professional user groups.
Previous approaches~\cite{saha2023dueling, zhao2023slic} follow the reinforcement learning from human feedback (RLHF)~\cite{ouyang2022training} paradigm: they first train a preference model to learn a neural-network-based reward function from human feedback, then fine-tune the target model with Proximal Policy Optimization (PPO)~\cite{schulman2017proximal} to maximize this reward.
Although RLHF has shown strong performance, the procedure is complex and unstable, with high computational overhead. Direct preference optimization (DPO)~\cite{rafailov2023direct} mitigates these drawbacks by eliminating explicit reward modeling and reinforcement learning, and directly optimizing the model using human preference data.
This approach has been successfully applied to improve LLMs performance in mathematical reasoning~\cite{wang2023math, luo2023wizardmath} and code generation~\cite{miao2024aligning, gee2024code}. We apply it to aesthetic image cropping by aligning the outputs of VLMs with human photography experts.

\section{Methodology}

\subsection{Overview}
The overall architecture of CROP is shown in~\cref{fig:model}.
We will introduce Compositional Reasoning Pipeline (CRP) in \cref{Perception} and Expert Preference Alignment (EPA) in \cref{Cognition} respectively.
\subsection{Compositional Reasoning Pipeline}
\label{Perception}

\begin{figure*}
	\centering
	\includegraphics[width=0.88\linewidth]{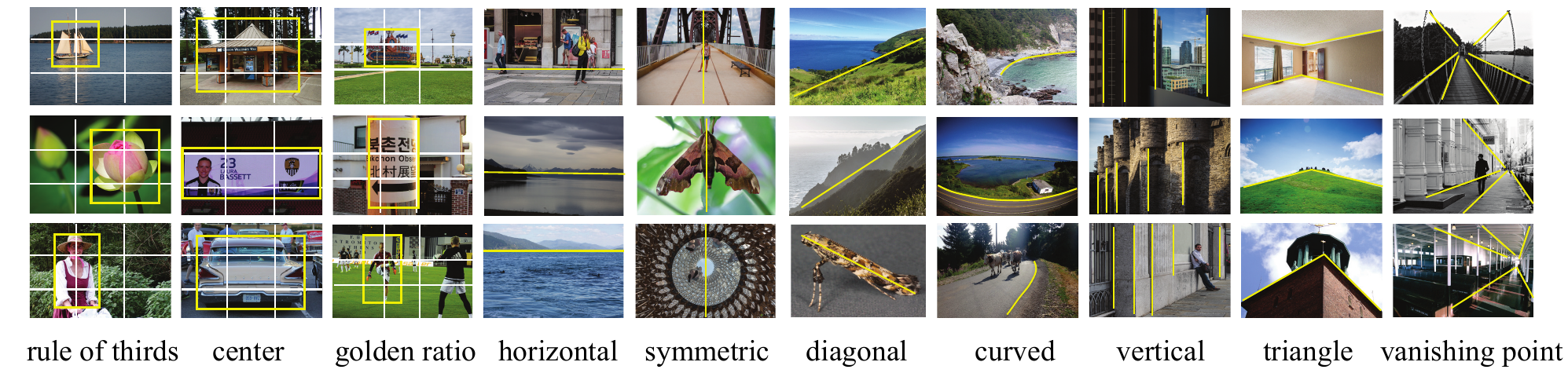}
    \caption{Diagram of ten important compositional elements. The photographer designs these elements to achieve a well-balanced photo layout. Image sourced from the CADB~\cite{zhang2021image} dataset.}
	\label{fig:rule}                                                                                             
\end{figure*}

We aim to align automatic aesthetic image cropping with professional, human-edited results. Accordingly, inspired by the practice of excellent photographers who analyze composition before pressing the shutter~\cite{freeman2017photographer}, we introduce compositional perception reasoning.

\textbf{Composition analysis.}
As illustrated in~\cref{fig:model} (a), we first feed the original image $I_\text{ori}$ into a visual encoder $E_\text{vis}$ to extract visual features. The VLM, denoted as $\Phi_\text{VLM}$, then conducts an initial reasoning pass with a specific prompt $P_\text{comp}$ to identify compositional elements. This process can be formulated as:
\begin{equation}\label{T}
T_\text{comp} = \Phi_\text{VLM}(E_\text{vis}(I_\text{ori}), P_\text{comp}),
\end{equation}
where $T_\text{comp} = \{(e_k, b_k)\}_{k=1}^{K}$ is the set of $K$ detected compositional elements, each consisting of a categorical label $e_k$ and positional coordinates $b_k$.
Following the classification of Lee et al.~\cite{lee2018photographic} and Zhang et al.~\cite{zhang2021image}, we focus on ten fundamental elements, namely: \emph{rule of thirds}, \emph{center},  \emph{golden ratio}, \emph{horizontal}, \emph{symmetric}, \emph{diagonal}, \emph{curved}, \emph{vertical}, \emph{triangle}, and \emph{vanishing point}, as shown in~\cref{fig:rule}.
The compositional information obtained from this analysis serves as prior knowledge for the VLM, which is then utilized in subsequent cropping proposal reasoning.

\textbf{Visual enhancement.}
\begin{table}[]
    \centering

       \caption{The normalized attention scores assigned by the VLM to different concept tokens in the prompt before generating the response. Higher scores indicate that the model places greater attention on that concept during inference.}
     \resizebox{0.30 \textwidth}{!}{\begin{tabular}{lcc}
    \toprule
     Elements  & Semantics $e_k$  & Coordinates $b_k$  \\
    \midrule
    Center&  \textbf{0.416} & 0.103 \\
    Vertical&  \textbf{0.390} & 0.024 \\
    Horizontal&  \textbf{0.047} & 0.041 \\
    \midrule
    Average &  \textbf{0.285} & 0.056 \\
    \bottomrule
    \end{tabular}} 
    \label{tab:attn}
\end{table}
Although the text analysis $T_\text{comp}$ provides both semantic labels $e_k$ and coordinates $b_k$, the spatial significance of the pure text coordinates $b_k$ is insufficient when used as input to the VLM, making it difficult to effectively constrain the subsequent reasoning process.
Taking the input image in~\cref{fig:model} as an example, we quantify the attention weights assigned by the VLM during inference to different tokens in $T_\text{comp}$.
As shown in~\cref{tab:attn}, the VLM places more attention on conceptual semantic tokens while largely ignoring spatial coordinate tokens. Therefore, we apply visual enhancement to amplify the prominence of spatial cues.
We define a visualization function $V(\cdot)$ that overlays graphical elements derived from $T_\text{comp}$ onto the original image $I_\text{ori}$, creating a visual enhancement image $I_\text{comp}$ for better compositional perception:
\begin{equation}\label{I}
I_\text{comp} = V(I_\text{ori}, T_\text{comp}).
\end{equation}

For composition elements that emphasize subject placement (e.g., rule of thirds, golden ratio), we overlay bounding boxes on the original image to highlight relative positions. For elements that emphasize global layout and alignment (e.g., vertical, horizontal), we draw guidelines to reveal the scene’s overall structure.

\textbf{Cropping proposal.}
Next, the visual enhancement image $I_\text{comp}$ is encoded and fused with the textual analysis $T_\text{comp}$ as the intermediate reasoning input to the VLM. Leveraging both visual and textual cues with a cropping prompt $P_\text{crop}$, the model conducts a second reasoning pass to predict candidate crops. This process is formulated as:
\begin{equation}\label{C}
\mathcal{C}_\text{cand} = \Phi_\text{VLM}(E_\text{vis}(I_\text{comp}), T_\text{comp}, P_\text{crop}),
\end{equation}
where $\mathcal{C}_\text{cand}=\{C_n\}_{n=1}^{N}$ is the set of $N$ candidate crops. These candidates are then forwarded to the downstream aesthetic decision module.\\
\textbf{Aesthetic decision.}
Finally, the VLM performs the aesthetic decision stage. It takes the set of candidate crops $\mathcal{C}_\text{cand}$ and the decision prompt $P_\text{aes}$ as input. The VLM must compare and evaluate the $N$ candidate crops from the perspectives of balance, focus, and visual appeal.
This selection process is formulated as:
\begin{equation}\label{D}
C_\text{final} = \Phi_\text{VLM}(\mathcal{C}_\text{cand}, P_\text{aes}),
\end{equation}
where $C_\text{final}$ is the selected crop that yields the best result.

The composition assessment database (CADB)~\cite{zhang2021image} provides a large number of composition elements annotated by professional experts. Based on this dataset, we construct a compositional reasoning dataset $D_\text{CRP}$. See supplementary materials for construction details.
Then we apply supervised fine-tuning with LoRA~\cite{hu2022lora} to enforce the output format and improve accuracy.
\subsection{Expert Preference Alignment}
\label{Cognition}
When facing such high-level semantic tasks, VLMs may exhibit hallucinations~\cite{leng2024mitigating, jiang2025multimodal} and other errors, leading to suboptimal or even poor outcomes.
To equip the VLM with complex decision-making abilities aligned with human experts, we design a two-stage expert cognition alignment process for aesthetic decision. Firstly, supervised fine-tuning (SFT) is used to instill fundamental aesthetic knowledge into the model. Then direct preference optimization (DPO) aligns its decision-making preference with those of human experts.

\textbf{Imitation learning.}
Firstly, we construct an aesthetic decision fine-tuning dataset, denoted as $D_\text{SFT}$, to preliminarily train the model’s aesthetic judgment ability.
Each sample in this dataset is a quadruple $(\mathcal{I}_i, \mathcal{C}_i, \mathcal{P}_i, y_i)$, where $\mathcal{I}_i$ represents the original input image, and $\mathcal{C}_i = \{C_n\}_{n=1}^{N}$ denotes a set of $N$ candidate cropping boxes corresponding to $\mathcal{I}_i$.
The GAICD~\cite{zeng2019reliable, zeng2020grid} dataset provides multiple crops annotated with mean opinion scores (MOS)~\cite{gao2022image} by experts. Based on the MOS distribution, we sample crops of varying quality to form $\mathcal{C}_i$.
Moreover, $\mathcal{P}_i$ serves as a guiding prompt, as shown in~\cref{fig:model} (b), which directs the model to perform the aesthetic decision-making task. $y_i$ denotes the target output, which is the cropping box with the highest MOS.

The goal of SFT is to train the $\Phi_\text{VLM}$ to imitate expert responses $y$. Let $\theta$ represent the trainable parameters of $\Phi_\text{VLM}$. The SFT objective is achieved by minimizing the following cross-entropy loss $L_\text{SFT}$ over the dataset $D_\text{SFT}$:
\begin{equation}\label{sft}
L_\text{SFT}(\theta) = - \mathbb{E}_{(x, y)} \left[ \sum_{t=1}^{|y|} \log \Phi(y_t | x, y_{<t}; \theta) \right],
\end{equation}
where $y_t$ denotes the $t$-th token of the response $y$, and $\Phi_\text{VLM}(y_t \mid \cdot)$ represents the model’s predicted probability of generating token $y_t$ given the input $x=(\mathcal{I}, \mathcal{C}, \mathcal{P})$ and the previously generated context $y_{<t}$.

\textbf{Preference alignment.}
The VLM fine-tuned with SFT alleviates hallucination issues and demonstrates stable aesthetic judgment capabilities. However, SFT assumes that each input $x$ has a unique response $y$, and trains the model to predict the correct response by maximizing $\Phi_\text{VLM}(y\mid x)$.
This results in the model not understanding why one answer is better than another. In contrast, human aesthetic decision-making is inherently a process of latent preference ranking. The absence of this process causes the VLM to often produce locally optimal results.

To overcome the limitations of SFT, we introduce DPO as a second-stage fine-tuning process to more precisely align the model with human expert preference.
We construct a preference dataset $D_\text{DPO}$, where the input $x = (\mathcal{I}, \mathcal{C}, \mathcal{P})$ is the same as in $D_\text{SFT}$, but the outputs consist of $y_\text{w}$ and $y_\text{l}$. $y_\text{w}$ represents the optimal cropping result, selected as the cropping box with the highest MOS from the GAICD dataset. $y_\text{l}$ represents the suboptimal result, which is derived from cropping boxes with relatively poorer performance.

The traditional reinforcement learning from human feedback (RLHF) process typically involves two complex stages: first, training an independent reward model (RM) on the preference dataset, and then using RM to optimize the policy through reinforcement learning.
In contrast, DPO proposes a more direct approach. It uses the model trained during the SFT phase (denoted as $\Phi_\text{ref}$) as the reference model and skips the step of explicitly training the RM.

The goal of DPO is to directly optimize the VLM policy $\Phi_{\theta}$ to maximize human preference in $D_\text{DPO}$. The loss function $L_\text{DPO}$ can be briefly expressed as:
\begin{small}
\begin{equation}\label{dpo}
L_\text{DPO}(\theta; \Phi_\text{ref}) = - \mathbb{E}_{(x, y_\text{w}, y_\text{l})}\\ \left[ \log \sigma \left( \hat{r}_{\theta}(x, y_\text{w}) - \hat{r}_{\theta}(x, y_\text{l}) \right) \right],
\end{equation}
\end{small}
where $\sigma$ is the sigmoid function. $\hat{r}_{\theta}(x, y)$ represents the implicit reward difference between the policy $\Phi_{\theta}$ and the reference policy $\Phi_\text{ref}$:
\begin{equation}\label{reward}
\hat{r}_{\theta}(x, y) = \beta \log \left( \frac{\Phi_{\theta}(y|x)}{\Phi_\text{ref}(y|x)} \right),
\end{equation}
where $\beta$ controls the extent of divergence from the reference model $\Phi_\text{ref}$.
$L_\text{DPO}$ encourages the model $\Phi_{\theta}$ to increase the relative probability assigned to the `winning' response $y_\text{w}$, while decreasing the relative probability assigned to the `losing' response $y_\text{l}$, thereby aligning the model's decision preference with those of human experts.

\section{Experiments}
\subsection{Implementation details}
\label{sec:detail}
\begin{table*}[t]
\caption{Comparison results with different technical approaches on three datasets. The best results are highlighted in bold, while the second-best results are underscored. “-” means that results are not available.}
	\centering
 \label{tab:main}
		\resizebox{0.85\linewidth}{!}{
			\begin{tabular}{lcccccccc}

				\toprule
				\multirow{2}{*}{Methods}&\multirow{2}{*}{Categories}&\multirow{2}{*}{Year}&\multicolumn{2}{c}{GAICD} &\multicolumn{2}{c}{FLMS}&\multicolumn{2}{c}{FCDB}\\
				\cmidrule(lr){4-5}\cmidrule(lr){6-7} \cmidrule(lr){8-9} 
				& & &ACC$_{1/5}$($\uparrow$) & \multicolumn{1}{c}{ACC$_{1/10}$($\uparrow$)} & IoU ($\uparrow$)&BDE($\downarrow$)& IoU ($\uparrow$)&BDE($\downarrow$)\\
                \midrule
                {Smartcrop~\cite{smartcrop}}&Hand-crafted features&2014&6.6&12.8&0.784&0.059&0.592&0.099\\
                \midrule
                {VFN~\cite{chen2017learning}}&\multirow{6}{*}{View ranking}&2017&27.0&39.0&0.577&0.124&0.685&0.084\\ 
                {A2-RL~\cite{li2018a2}}&&2018&23.2&39.5&0.821&0.045&0.664&0.089\\
                {VPN~\cite{wei2018good}}&&2018&40.0&49.5&0.835&\underline{0.033}&0.711&0.073\\ 
                {VEN~\cite{wei2018good}}&&2018&40.5&54.0&0.837&0.041&\underline{0.735}&0.072\\ 
                {CGS~\cite{li2020composing}}&&2020&63.0&81.5&0.836&0.039&0.685&0.079\\ 
                {GAIC~\cite{zeng2020grid}}&&2020&68.2&84.4&0.834&0.041&0.674&0.081\\ 
                 \midrule
                {CACNet~\cite{hong2021composing}}&\multirow{4}{*}{Saliency prediction}&2021&68.8&85.4&0.843&0.036&0.694&0.076\\ 
                {Jia \textit{et al}~\cite{jia2022rethinking}}&&2022&81.5&91.0&0.838&0.037&-&-\\ 
                {Wang \textit{et al}~\cite{wang2023image}}&&2023&70.0&86.8&-&-&0.695&0.075\\
                {CAGR~\cite{zhu2025image}}&&2025&-&-&\underline{0.851}&\underline{0.033}&0.721&\underline{0.067}\\ 
                 \midrule
                {ProCrop~\cite{zhang2025procrop}}&Retrieval-based&2025&\underline{85.4}&\underline{94.2}&0.843&0.036&-&-\\
                {Cropper~\cite{lee2025cropper}}&VLM + Retrieval-based&2025&73.4&90.6&0.818&0.045&0.655&0.090\\ 
                 \midrule
                GPT-5~\cite{gpt5}&\multirow{2}{*}{VLM}&2025&26.9&37.8&0.798&0.046&0.606&0.102\\
                Ours&&2025&\textbf{86.2}&\textbf{94.8}&\textbf{0.871}&\textbf{0.027}&\textbf{0.745}&\textbf{0.056}\\
                \bottomrule
		      \end{tabular}}
        	
\end{table*}

\textbf{Datasets.}
The Composition Assessment Database (CADB) \cite{zhang2021image} contains 9,497 images with each image rated by 5 individual raters who specialize in fine art for the overall composition quality.
For each image, CADB provides a wealth of annotated compositional elements, as shown in~\cref{fig:rule}. We use these compositional elements as an intermediate step, constructing 9,204 dialogue samples for training in the composition analysis and cropping proposal stages.
GAICD~\cite{zeng2020grid} dataset has 3,336 images, with 2,636 for training, 200 for validation, and 500 for testing, containing 288,069 densely annotated crops with mean opinion scores (MOS).
We use these crops to construct 4,000 supervised fine-tuning samples and 8,000 preference samples for training in the aesthetic decision stage.
See the Appendix for more details.

In addition to performing inference evaluations on the GAICD~\cite{zeng2020grid} test set, following~\cite{lee2025cropper, zhang2025procrop, zhu2025image}, we also test on the FCDB~\cite{chen2017quantitative} (3,414 images, with 345 images in the test set) and FLMS~\cite{FLMS_mm14} (500 images) datasets to validate the generalization.

\textbf{Training and inference.} 
We adopt Qwen2.5-VL-7B~\cite{bai2025qwen2} as the base model, with the maximum sequence length set to 4096 tokens. All experiments are conducted under the ms-swift~\cite{ms-swift} training framework using a two-stage LoRA~\cite{hu2022lora} fine-tuning strategy, where the parameters are configured as $r$ = 16 and $\alpha$ = 32. The ViT~\cite{dosovitskiy2020image} backbone is frozen.
All experiments run on a single RTX 4090 GPU (48 GB VRAM).
Both stages use an effective batch size of 16, 5\% warm-up ratio, and 3 training epochs.
SFT uses a learning rate of $1\times10^{-4}$ and the loss function follows~\cref{sft}.
DPO reduces the learning rate to $1\times10^{-5}$ with coefficient $\beta$ is set to 0.2, and the loss function is defined as in~\cref{dpo}.
During inference, we set the temperature to 0.1 and top-p to 0.95.

\textbf{Metrics.}
We use standard metrics in the image cropping, including the ACC$_{K/N}$, Intersection over Union (IoU) and Boundary Displacement Error (BDE).
Specifically, ACC$_{K/N}$ indicates whether top-$K$ from predictions could be involved among top-$N$ crops from
the ground-truth based on MOS.
When predicted views do not align exactly with predefined grid views, we consider two crops equivalent if their IoU exceeds a threshold of $\epsilon$ = 0.85, as in~\cite{zhang2025procrop, jia2022rethinking, liu2023beyond}.
IoU is defined as the ratio of the intersection area to the union area of the predicted and ground-truth crop bounding box.
BDE represents the average $L_1$ distance between the ground-truth coordinates and the predicted values.
We also conduct a user study for subjective evaluation in~\cref{user}.
\subsection{Comparative Evaluation}
\begin{figure*}
	\centering
	\includegraphics[width=0.89\linewidth]{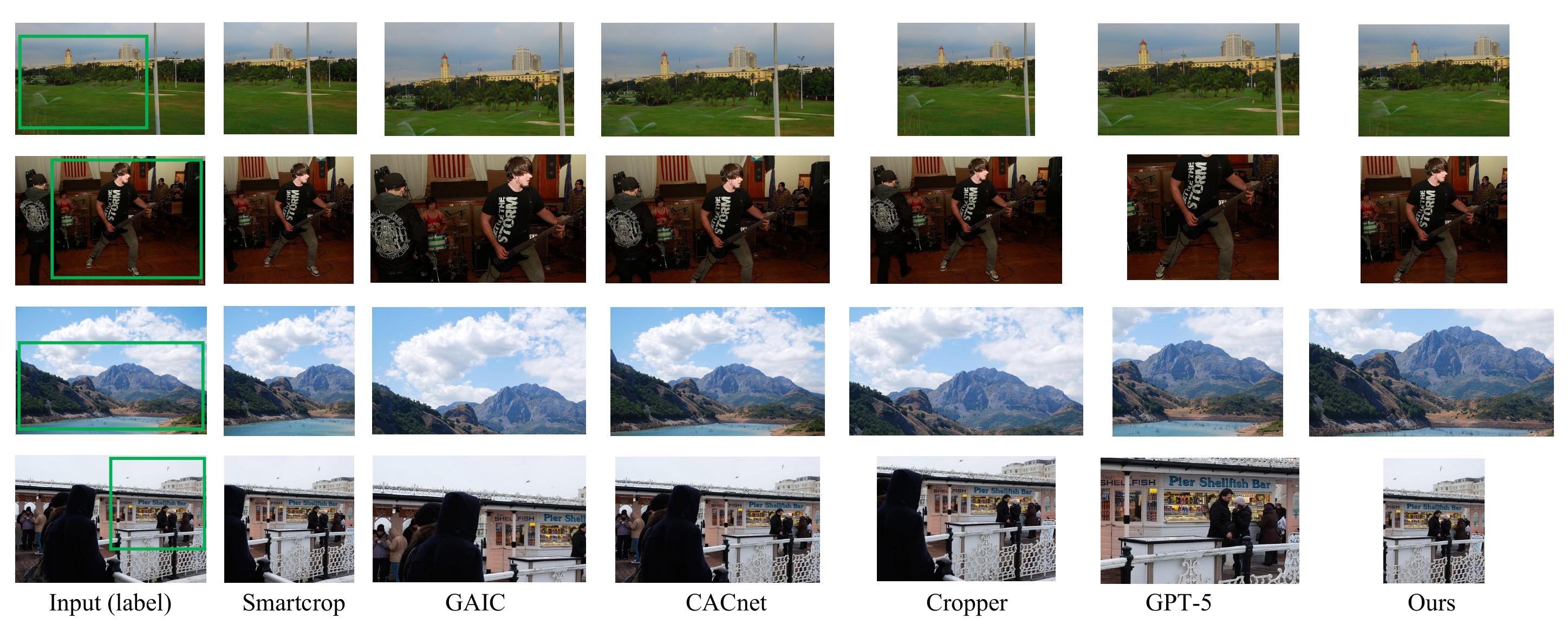}
    \caption{Visual comparison to other methods. The labels are from the FCDB~\cite{chen2017quantitative} dataset.}
	\label{fig:compare}                                                                                             
\end{figure*}
We compare our method with approaches based on different technical approaches.
Specifically, Smartcrop~\cite{smartcrop} represents traditional methods relying on hand-crafted features, GAIC~\cite{zeng2020grid} and others~\cite{li2018a2, wei2018good, zeng2020grid, li2020composing} represent view-ranking-based methods, CACNet~\cite{hong2021composing} and similar methods~\cite{hong2021composing, jia2022rethinking, wang2023image} represent popular deep learning approaches based on saliency prediction, while Cropper~\cite{lee2025cropper} and ProCrop~\cite{zhang2025procrop} represent new methods based on retrieval enhancement.

\cref{tab:main} shows the comparison results of our method and the methods mentioned above on three datasets.
FCDB~\cite{chen2017quantitative} and FLMS~\cite{FLMS_mm14} primarily evaluate the localization accuracy of the cropping box generated by the method.
On the other hand, GAICD~\cite{zeng2020grid} emphasizes the alignment accuracy between the generated cropping boxes and human aesthetic preference.
\cref{fig:compare} shows the visualization results of our method and other representative methods.

Due to its reliance on hand-crafted features such as edges and faces, Smartcrop is constrained by rigid rules and performs poorly on the metric and visual effects.
View-ranking-based methods achieve good performance on the FLMS and FCDB datasets. However, these methods make it difficult to align their outputs with human aesthetic preference on the GAICD dataset, resulting in a poor visual effect.
In contrast, saliency-based methods benefit from saliency-weighted composition, making their results more consistent with human visual judgments and improving performance on the GAICD dataset. However, the mediocre cropping results in these methods still lack aesthetic appeal.
As shown in~\cref{fig:compare}, CACNet does almost no effective cropping of the original image.

For retrieval-based methods, although Cropper leverages a VLM to predict the crops, its reliance on inaccurate cosine similarity retrieval features resulting in weak visual performance. ProCrop improves upon this by retrieving compositional features from database, achieving the second-best results on the GAICD dataset. However, the positioning of the cropping box is still not accurate enough.
Meanwhile, the zero-shot performance of leading commercial models remains unsatisfactory.
Our method precisely localizes and analyzes compositional elements through the CR, achieving the best performance on objective metrics. The EPA training further aligns our cropping results with those of human experts, producing superior visual outcomes under different scenes, as shown in~\cref{fig:compare}.
\subsection{Ablation Study}
\begin{table}[]

	\centering
         \caption{
    Ablation analysis on compositional reasoning pipeline (CRP), visual enhancement (VE), and expert preference alignment (EPA) modules. Best results are shown in bold, and second-best results are underlined.}
    \label{tab:ablation}
		\resizebox{0.95\linewidth}{!}{
			\begin{tabular}{cccccccc}

				\toprule
				\multirow{2}{*}{ID}&\multicolumn{3}{c}{Modules}&\multicolumn{2}{c}{GAICD}&\multicolumn{2}{c}{FLMS}\\
				\cmidrule(lr){2-4}\cmidrule(lr){5-6} \cmidrule(lr){7-8} 
				&CRP&VE&EPA&\multicolumn{1}{c}{ACC$_{1/5}$($\uparrow$)}&\multicolumn{1}{c}{ACC$_{1/10}$($\uparrow$)}& IoU ($\uparrow$)&BDE($\downarrow$)\\
                \midrule
                
                C1&\ding{55}&\ding{55}&\ding{55}&76.0&89.2&0.822&0.036\\
                C2&\ding{55}&\ding{55}&\ding{51}&80.8&91.4&0.852&0.030\\
                C3&\ding{51}&\ding{55}&\ding{55}&81.4&91.6&0.852&0.032\\
                C4&\ding{51}&\ding{51}&\ding{55}&82.0&92.0&0.858&0.030\\
                C5&\ding{51}&\ding{55}&\ding{51}&\underline{85.5}&\underline{94.2}&\underline{0.867}&\underline{0.028}\\
                C6&\ding{51}&\ding{51}&\ding{51}&\textbf{86.2}&\textbf{94.8}&\textbf{0.871}&\textbf{0.027}\\
                \bottomrule
		      \end{tabular}}

\end{table}

\begin{figure}
	\centering
	\includegraphics[width=0.90\linewidth]{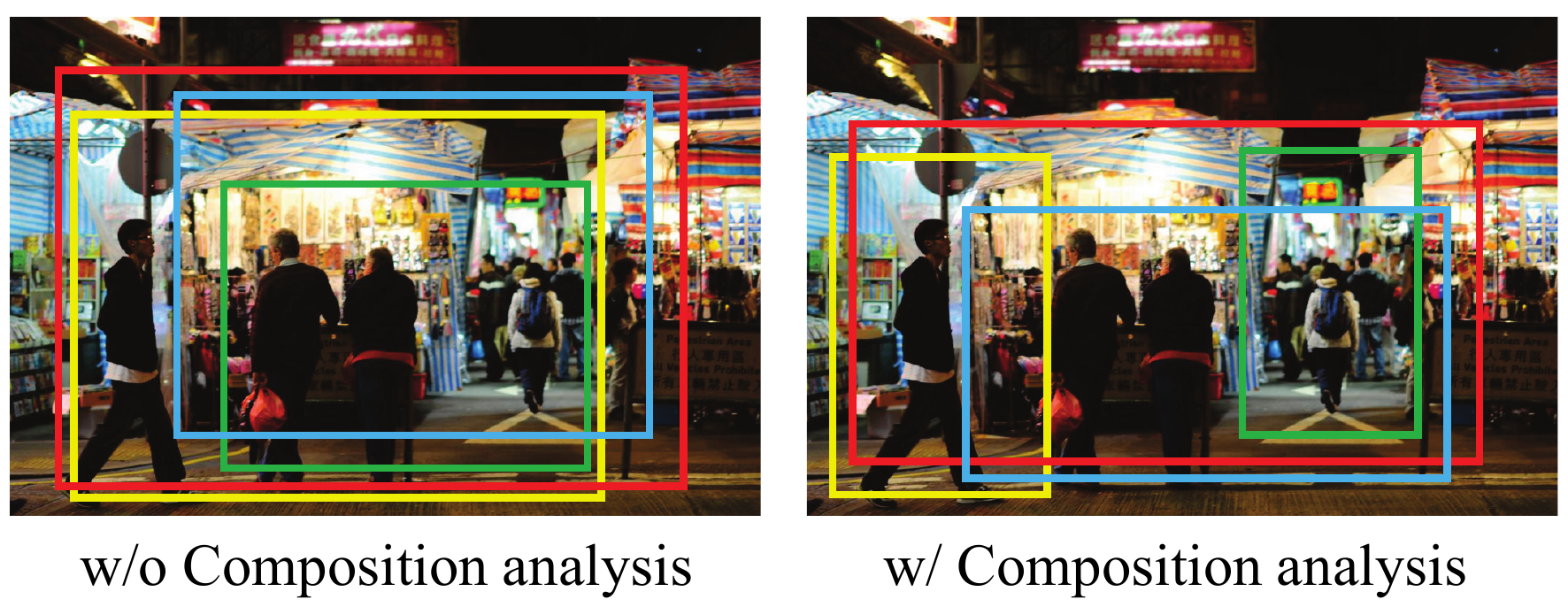}
    \caption{The impact of composition analysis on candidate crops.}
	\label{fig:ca}                                                                                             
\end{figure}
\begin{figure}
	\centering
	\includegraphics[width=0.95\linewidth]{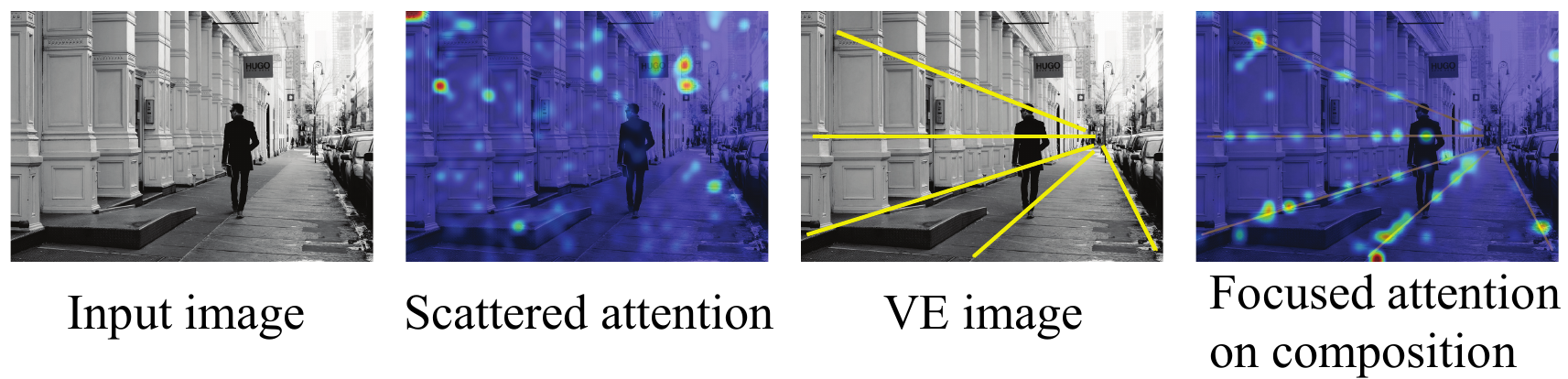}
    \caption{Visualization of cross-modal attention from compositional keywords to input image.}
	\label{fig:ve}                                                                                             
\end{figure}
\begin{figure}
	\centering
	\includegraphics[width=0.90\linewidth]{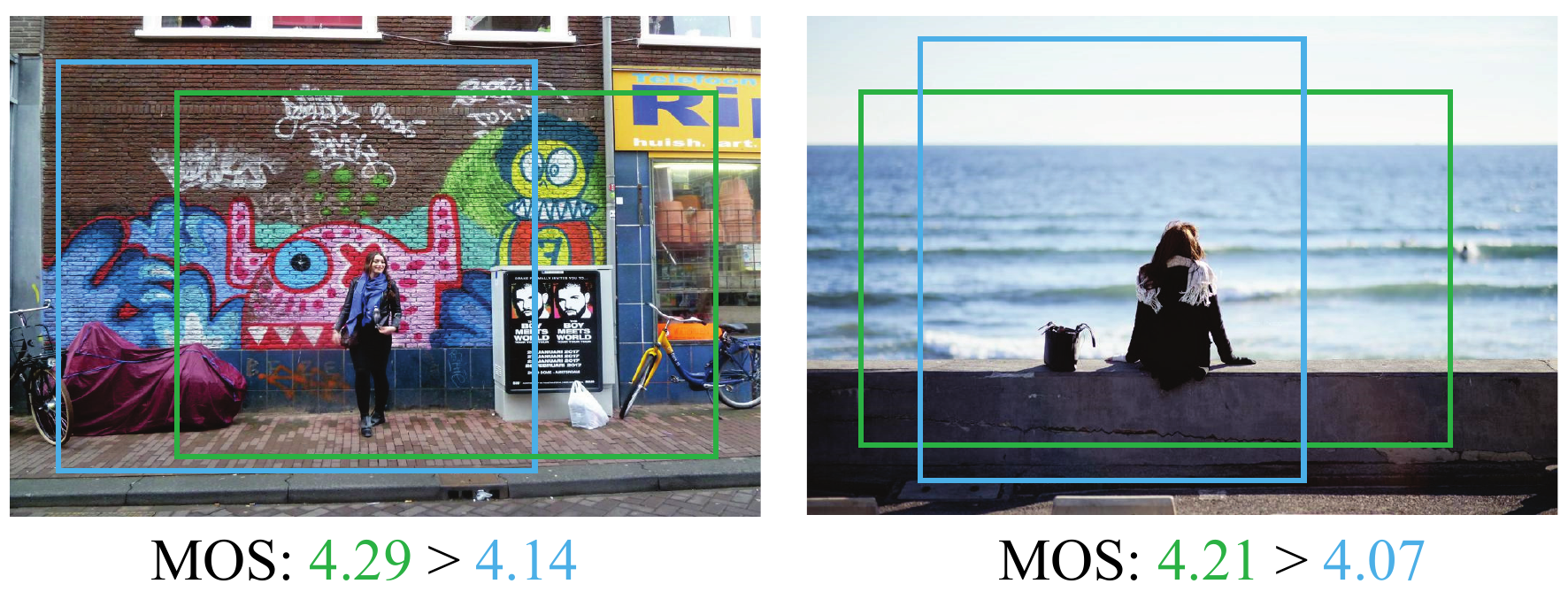}
    \caption{The impact of DPO on aesthetic decisions: \textcolor{myCustomGreen}{green} represents the choice of the DPO model, and \textcolor{myCustomBlue}{blue} represents the choice of the SFT model. }
	\label{fig:dpo}                                                                                             
\end{figure}
\begin{figure*}
	\centering
	\includegraphics[width=0.90\linewidth]{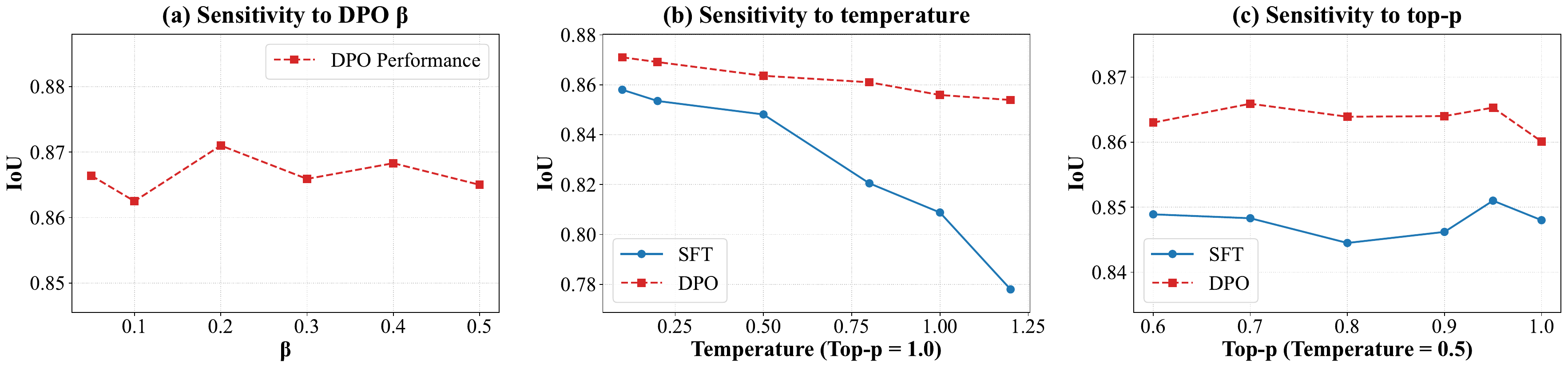}
    \caption{Sensitivity to training hyperparameter $\beta$, inference hyperparameter temperature and top-p on the FLMS~\cite{FLMS_mm14}.}
	\label{fig:sens}                                                                                             
\end{figure*}
In this section, we conduct comprehensive ablation studies to validate the effectiveness of each component in our proposed model. As shown in~\cref{tab:ablation}, we numbered the different ablation combinations.

\textbf{Effect of compositional reasoning pipeline.}
C1 represents a baseline model which performs simple coordinate regression using the VLM with a single-stage fine-tuning process.
Through step-by-step reasoning about compositional elements and rules, C3 achieves significant improvements across all metrics. As shown in~\cref{fig:ca}, the explicit compositional analysis stage makes VLM's cropping proposals more specific and clear.

\textbf{Effect of visual enhancement.}
Visual enhancement strengthens the model’s vision–language interaction. Quantitatively, adding VE (C4 and C6) brings a stable performance improvement. We also visualize cross-modal attention from compositional keywords (e.g., “vanishing point”) to the input image in~\cref{fig:ve}, showing that visual enhancement biases the VLM toward compositional structure rather than incidental content.

\textbf{Effect of expert preference alignment.}
Through direct preference optimization, expert preference alignment enhances the model’s aesthetic judgment. C2 and C5 respectively improve the performance of the baseline and CRP models. As illustrated in~\cref{fig:dpo}, when given candidate crops with different MOS scores during the decision stage, the DPO model consistently selects higher-quality results compared with SFT.
By integrating all components, our full model C6 achieves the best overall performance, demonstrating the collaborative capabilities between these modules.

\textbf{Effect of fine-tuning.}
We further evaluated the performance of the zero-shot model without fine-tuning, as shown in~\cref{tab:fintune}. Although Qwen3-VL-plus~\cite{Qwen3} and GPT-5~\cite{gpt5} are state-of-the-art closed-source VLMs, their performance is still inferior to that of the fine-tuned 7B model.
In particular, on the GAICD~\cite{zeng2020grid} dataset, where the IoU threshold $\epsilon$ follows the setting in~\cite{zhang2025procrop, jia2022rethinking, liu2023beyond} and is set to 0.85, most of the cropping boxes generated by these models fall below this threshold, resulting in extremely low ACC$_{1/5}$ and ACC$_{1/10}$ scores.
Therefore, we argue that fine-tuning VLMs is essential for aesthetic image cropping.
\begin{table}[!htb]

	\centering
         \caption{Ablation study on fine-tuning strategies.}
    \label{tab:fintune}
		\resizebox{0.99\linewidth}{!}{
			\begin{tabular}{lccccc}

				\toprule
				\multirow{2}{*}{Models}&\multirow{2}{*}{Fine-tuning}&\multicolumn{2}{c}{GAICD}&\multicolumn{2}{c}{FLMS}\\
				\cmidrule(lr){3-4} \cmidrule(lr){5-6} 
			&&\multicolumn{1}{c}{ACC$_{1/5}$($\uparrow$)}&\multicolumn{1}{c}{ACC$_{1/10}$($\uparrow$)}& IoU ($\uparrow$)&BDE($\downarrow$)\\
                \midrule
                Qwen2.5-VL-7B~\cite{bai2025qwen2}&zero-shot&1.8&2.4&0.621&0.078\\
                Qwen3-VL-plus~\cite{Qwen3}&zero-shot&5.1&10.3&0.714&0.074\\
                GPT-5~\cite{gpt5}&zero-shot&26.9&37.8&0.798&0.046\\
                \midrule
                Ours (Qwen2.5-VL-7B)&SFT&76.0&89.2&0.822&0.036\\
                Ours (Qwen2.5-VL-7B)&SFT+DPO&\textbf{86.2}&\textbf{94.8}&\textbf{0.871}&\textbf{0.027}\\
                \bottomrule
		      \end{tabular}}

\end{table}
\subsection{Parameter sensitivity analysis}
As shown in~\cref{reward}, $\beta$ controls the balance between the reward signal and the generative model, thus affecting the quality and diversity of outputs.
In~\cref{fig:sens} (a), the model was trained with $\beta$ ranging from 0.05 to 0.5. Smaller $\beta$ values yield conservative outputs and weaken preference learning, while larger ones promote diversity but may cause instability. The best performance is achieved at $\beta=0.2$.

In the inference stage, temperature controls the randomness of the generated text. Lower temperature values make the output more deterministic, while higher temperatures increase randomness.
\cref{fig:sens} (b) illustrates the performance differences between SFT and DPO as temperature varies from 0.1 to 1.2. As temperature increases, the uncertainty leads to a decline in performance for both SFT and DPO stages. Notably, the performance drop in SFT is more obvious, whereas DPO significantly enhances the stability.

The $\text{top-p}\in(0,1]$ regulates diversity by limiting candidate tokens based on cumulative probability. Since its effect is weakened at low temperature, we fix the temperature at 0.5 and vary top-p from 0.6 to 1.0. As shown in~\cref{fig:sens} (c), both models perform best at $\text{top-p}=0.95$, with DPO consistently maintaining superior and stable results.

\subsection{User Study}
To validate the practical effectiveness, we conducted a user study with 10 participants.
We randomly sampled 150 test images from GAICD~\cite{zeng2020grid}, FCDB~\cite{chen2017quantitative}, and FLMS~\cite{FLMS_mm14}, creating pairwise comparisons against GAIC~\cite{zeng2020grid}, CACNet~\cite{hong2021composing}, and Cropper~\cite{lee2025cropper}.
Participants were asked to choose the more aesthetically appealing crop. As shown in~\cref{tab:user_study}, our method was significantly preferred across the 1,500 collected votes, confirming its superior visual appeal.
See the Appendix for more details.
\label{user}
\begin{table}[]
    \centering
       \caption{User study: Preference rate v.s. GAIC~\cite{zeng2020grid}, CACNet~\cite{hong2021composing}, and Cropper~\cite{lee2025cropper}.}
     \resizebox{0.37\textwidth}{!}{\begin{tabular}{lcc}
    \toprule
    Choice  & Baseline ($\%$) & CROP ($\%$) \\
    \midrule
    GAIC~\cite{zeng2020grid} v.s. CROP  &  20.8 & \textbf{79.2} \\
    CACNet~\cite{hong2021composing} v.s. CROP  &  38.0 & \textbf{62.0} \\
    Cropper~\cite{lee2025cropper} v.s. CROP  &  32.0 & \textbf{68.0} \\
    \bottomrule
    \end{tabular}}
    \label{tab:user_study}
\end{table}
\section{Conclusion and Limitation}
In this paper, we propose an aesthetic image cropping method based on compositional reasoning pipeline and expert preference alignment. The proposed approach fully exploits the VLM’s capability for aesthetic understanding and reasoning. Comprehensive experiments demonstrate the effectiveness and superiority of our method.
In the future, we plan to further explore VLMs for user-oriented and personalized image cropping. The main limitation of the current approach lies in its computational cost. As lightweight VLMs continue to advance, they will accelerate deployment and enable broader real-world applications.

\section*{Impact Statement}

This paper presents work whose goal is to advance the field of Machine Learning. Specifically, we focus on exploring the multimodal reasoning capabilities of VLMs in the domain of computational aesthetics, aiming to bridge the gap between automated image cropping and professional human composition standards through explicit reasoning and preference alignment. There are many potential societal consequences of our work, none which we feel must be specifically highlighted here.

\bibliography{icml2026}
\bibliographystyle{icml2026}

\clearpage
\appendix
\begin{center}
\Large\textbf{Appendix}
\label{appendix}
\end{center}
\section{Complete Prompts}
\label{supp:prompt}
\begin{table}[!htb]
    \centering
       \caption{Prompts for VLM.}
     \begin{tabularx}{\columnwidth}{l Y}
    \toprule
    \textbf{Prompt}&\textbf{Instruction}\\
    \midrule
    Baseline&``Analyze the image from the perspective of aesthetics and composition, then crop it to achieve the best visual effect. Please generate only a Python list containing four numbers representing the crop box coordinates, in the format [x1, y1, x2, y2]."\\
    Analysis Prompt&``You are a professional photography composition expert. First, analyze the photo and identify the key composition elements. Return them in JSON format, where each element includes its category and bounding box coordinates."\\
    Proposal Prompt&``Second, based on the detected composition elements, analyze which photography composition rules are most suitable for this image. Using those rules, generate several aesthetically pleasing cropping boxes, and return them as a Python list in the format [[x1, y1, x2, y2], ...]."\\
    Decision Prompt&``Finally, your task is to act as an aesthetic decider. Critically evaluate all the candidate crops based on their visual appeal, balance, and how well they focus on the subject. Select the letter ID of the single best crop that results in the most beautiful and compelling image."\\
    \bottomrule
    \end{tabularx}
    \label{tab:prompt}
\end{table}
We present the complete VLM prompts in~\cref{tab:prompt}. In the ablation study, Baseline refers to single-step inference without CRP, where the model directly outputs the final result. The Analysis Prompt, Proposal Prompt, and Decision Prompt correspond to the three stages of CROP, respectively.
Notably, before the decision stage, all candidate crops are assigned unique IDs, and the model produces the final output by selecting the corresponding ID. This strategy effectively mitigates rare hallucination issues that may occur during VLM inference.
\section{Details of Datasets}
\label{sec:data}
In this section, we introduce the details of constructing the VLM dialogue training datasets, including the compositional reasoning dataset $D_\text{CRP}$ and the preference optimization dataset $D_\text{DPO}$.
\subsection{Compositional Reasoning Dataset}
To enable intermediate supervision for the Compositional Reasoning Pipeline, we use the composition assessment database (CADB)~\cite{zhang2021image} as the source of annotations. The CADB dataset provides a large number of composition elements, with a total of 13 types annotated by professional experts.
We exclude a few elements that have minimal impact on composition and select 10 key compositional elements, as illustrated in~\cref{fig:rule}.

The pretrained Qwen2.5-VL-7B~\cite{bai2025qwen2} model is capable of effectively handling JSON-structured content. Leveraging this feature, we extract the compositional element category labels and their corresponding bounding box coordinates from the CADB dataset to construct JSON-formatted response data, which are used for dialogue fine-tuning in the composition analysis stage.
Each dialogue sample consists of an input image, an analysis prompt, and a JSON-formatted response. The response includes one or more compositional elements, and each element corresponds to one or more bounding boxes, depending on its type.

In the cropping proposal stage, we use the dialogue from the composition analysis stage as contextual information. We then follow the definitions of compositional rules in~\cite{freeman2017photographer, prakel2020composition, zakia2012photographic} to generate candidate cropping boxes based on the identified compositional elements.
Consistent with the element categorization in the CADB dataset, for composition elements that emphasize subject placement (e.g., rule of thirds, golden ratio), we constrain the subject to the corresponding position within the cropping box. For elements that emphasize global layout and alignment (e.g., vertical, horizontal), we ensure that the cropping box encompasses these structural cues.
\begin{figure*}[!htb]
    \centering
    \includegraphics[width=0.8\linewidth]{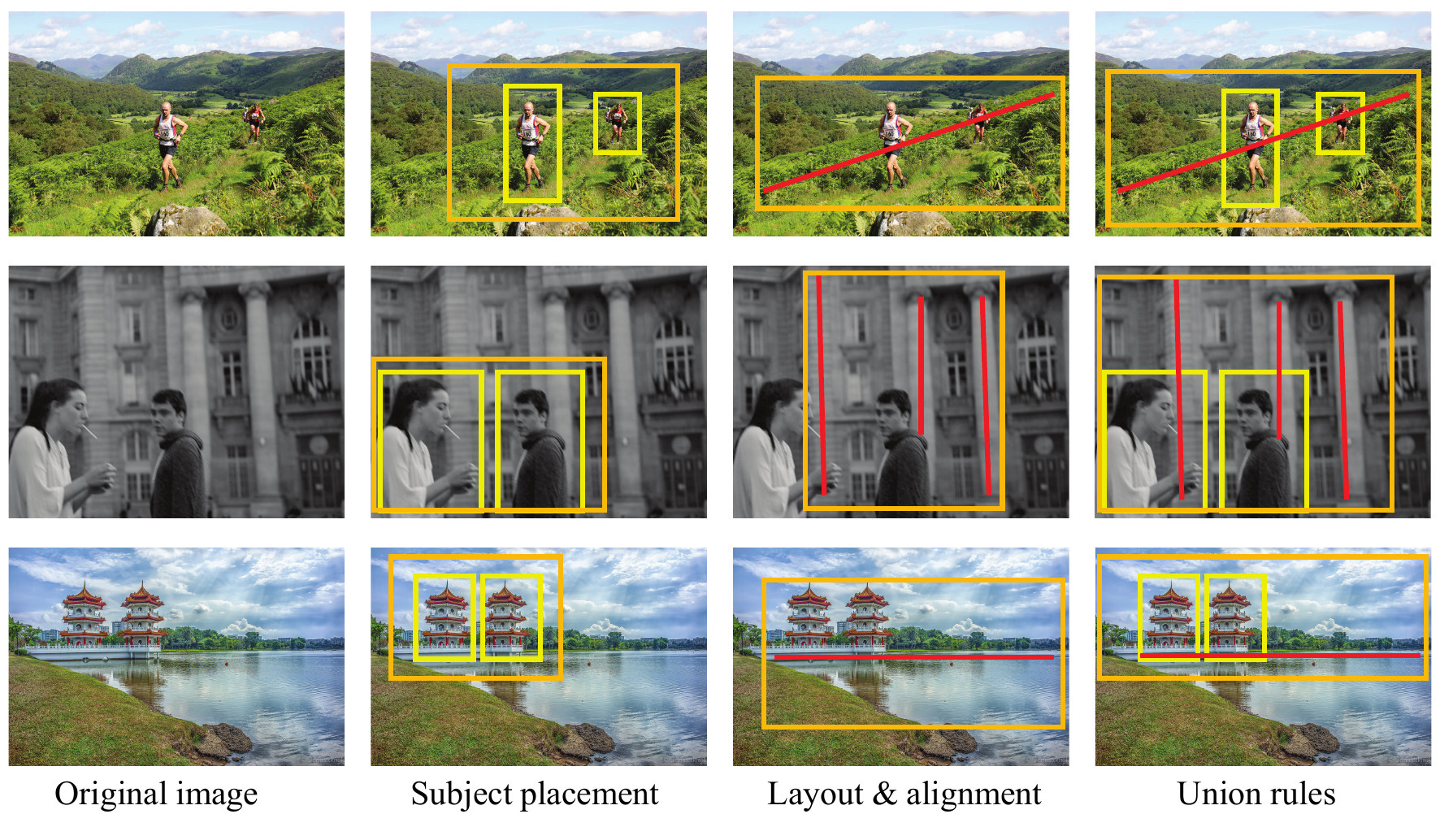}
    \caption{Proposal examples based on different types of composition rules.}
	\label{fig:data}  
\end{figure*}
For images that contain both of the above compositional elements, we add an additional cropping box to satisfy both rules. Examples are shown in the~\cref{fig:data}.
For images containing fewer compositional elements, we introduce random perturbations to existing cropping boxes to expand the set of candidate crops to a total of eight.

The aesthetic decision stage is independent of the previous two stages. We use the GAICD~\cite{zeng2020grid} dataset as the source of annotations.
GAICD dataset has 3,336 images, with 2,636 for training, 200 for validation, and 500 for testing, containing 288,069 densely annotated crops with mean opinion scores (MOS)~\cite{gao2022image}.
For each image in the training set, the MOS ranges from 1 to 5, with higher scores indicating better quality. We define a good crop threshold of 3.5 and a poor crop threshold of 2.5, while crops with scores between 2.5 and 3.5 are considered average.
Based on the distribution of the original MOS, we randomly select eight candidate cropping boxes from these intervals as the input candidates, and use the crop with the highest MOS score as the output target. Each cropping box is indexed and paired with a decision prompt to form a dialogue sample.
In total, we construct $D_\text{CRP}$, including 9,204 dialogue samples for training in the composition analysis and cropping proposal stages, 4,000 dialogue samples for the aesthetic decision stage.
\subsection{Preference Optimization Dataset}
To align the VLM with expert aesthetic cognition, we further construct a direct preference optimization dataset $D_\text{DPO}$, based on the supervised fine-tuning dataset $D_\text{SFT}$ from the aesthetic decision stage.
For the construction of input candidate crops, we follow the same setup as in $D_\text{SFT}$. The dialogue outputs are divided into a chosen response and a rejected response. Specifically, the cropping box with the highest MOS score is selected as the chosen response, while those with lower MOS scores are used as rejected responses.
It is worth noting that the rejected responses are not only selected from crops with MOS scores below 3.5, but also include suboptimal samples to further enhance the VLM’s judgment capability.
Finally, we construct 8,000 preference pairs for DPO training.
\section{Details of Training}
Based on the datasets constructed in~\cref{sec:data}, we train the Qwen2.5-VL-7B~\cite{bai2025qwen2} model. Since the three stages of the compositional reasoning pipeline have different output objectives, we adopt a stage-wise training strategy to reduce training complexity. Following the parameter settings in~\cref{sec:detail}, we first perform LoRA-based supervised fine-tuning for the composition analysis stage, then train the cropping proposal stage with contextual inputs. Finally, we separately train the aesthetic decision stage model. This stage-wise approach produces three sets of LoRA weights, which are switched accordingly during inference to reduce VRAM consumption.
For expert cognition alignment, we conduct additional DPO training on top of the SFT model in the aesthetic decision stage.
\section{Details of User Study}
For our user study, we recruited 10 participants, all of whom are graduate students specializing in computer vision.
Prior to the evaluation, participants received a briefing session that covered the task's definition, objectives, and evaluation criteria. Participants were instructed to select the more visually appealing crop based on aesthetic considerations. The evaluation interface is shown in~\cref{fig:user}. To eliminate bias, all methods were presented anonymously. The entire study took approximately 30 minutes for each participant. Finally, the overall preference for each method was calculated by aggregating the results from all participants.
\begin{figure}[!htb]
    \centering
    \includegraphics[width=0.98\linewidth]{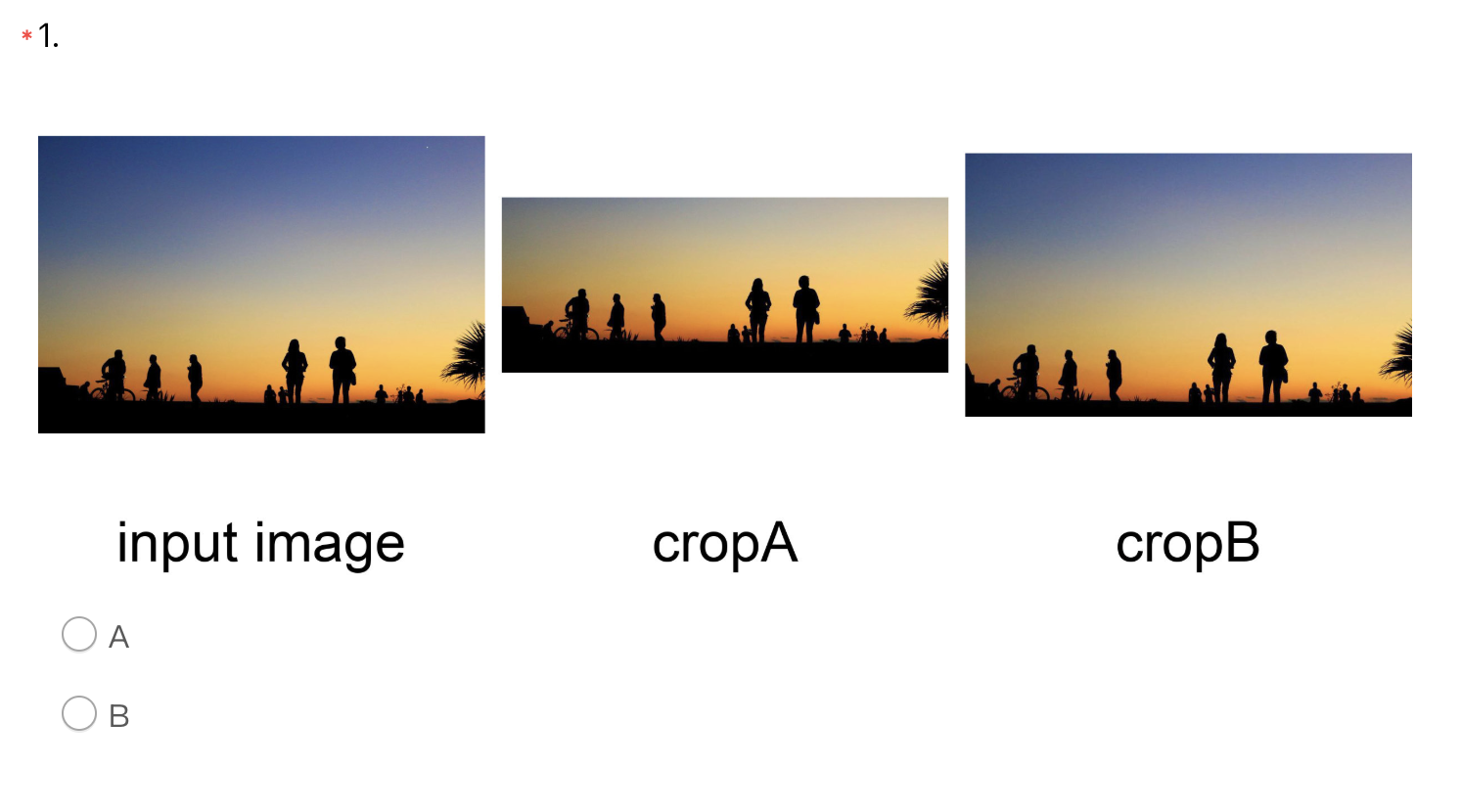}
    \caption{Evaluation interface for user study.}
	\label{fig:user}  
\end{figure}
\section{Failure Case Analysis}
We observe that the composition analysis step may struggle with dense visual clutter or significantly zoomed-in content, which compromises the final cropping quality. Future improvements could involve integrating a robust visual encoder to support better fine-grained perception.
\begin{figure}[!htb]
    \centering
    \includegraphics[width=0.98\linewidth]{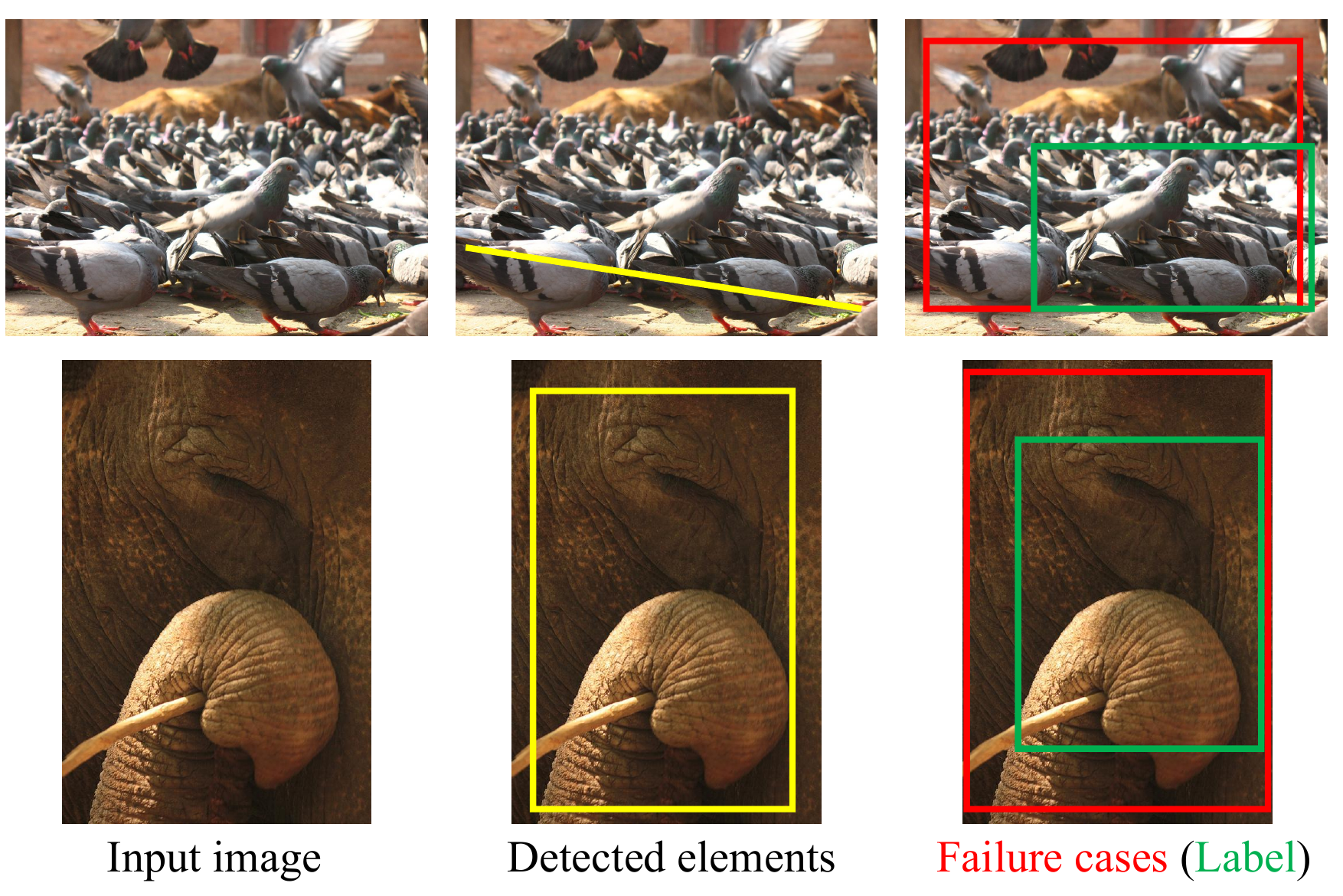}
    \caption{Failure cases.}
	\label{fig:user}  
\end{figure}

\end{document}